%% file: report.tex
\documentclass[10pt, letterpaper]{article}

\usepackage{graphicx} % more modern
\usepackage{amsmath,amssymb,amsthm}

\usepackage{booktabs}

\usepackage{color,mathtools}
\usepackage[round]{natbib}

\usepackage{multirow}
\usepackage{enumitem}

\usepackage{algorithm}
\usepackage{algorithmic}

\let\cite\citep

\usepackage[hyperindex,breaklinks]{hyperref}

\input{epfml_macros.tex}
\usepackage{fullpage}

\usepackage{cleveref}

\makeatletter
\newcommand\Autoref[1]{\@first@ref#1,@}
\def\@throw@dot#1.#2@{#1}% discard everything after the dot
\def\@set@refname#1{%    % set \@refname to autoefname+s using \getrefbykeydefault
    \edef\@tmp{\getrefbykeydefault{#1}{anchor}{}}%
    \xdef\@tmp{\expandafter\@throw@dot\@tmp.@}%
    \ltx@IfUndefined{\@tmp autorefnameplural}%
         {\def\@refname{\@nameuse{\@tmp autorefname}s}}%
         {\def\@refname{\@nameuse{\@tmp autorefnameplural}}}%
}
\def\@first@ref#1,#2{%
  \ifx#2@\autoref{#1}\let\@nextref\@gobble% only one ref, revert to normal \autoref
  \else%
    \@set@refname{#1}%  set \@refname to autoref name
    \@refname~\ref{#1}% add autoefname and first reference
    \let\@nextref\@next@ref% push processing to \@next@ref
  \fi%
  \@nextref#2%
}
\def\@next@ref#1,#2{%
   \ifx#2@ and~\ref{#1}\let\@nextref\@gobble% at end: print and+\ref and stop
   \else, \ref{#1}% print  ,+\ref and continue
   \fi%
   \@nextref#2%
}

\makeatother

\commenter{seb}

\title{Affine-Invariant Robust Training}
\author{Oriol Barbany Mayor}
\date{}

\begin{document}
\maketitle

\begin{abstract} 
The field of adversarial robustness has attracted significant attention in machine learning. Contrary to the common approach of training models that are accurate in average case, it aims at training models that are accurate for worst case inputs, hence it yields more robust and reliable models. Put differently, it tries to prevent an adversary from fooling a model. The study of adversarial robustness is largely focused on $\ell_p-$bounded adversarial perturbations, i.e.\ modifications of the inputs, bounded in some $\ell_p$ norm. Nevertheless, it has been shown that state-of-the-art models are also vulnerable to other more natural perturbations such as affine transformations, which were already considered in machine learning within data augmentation. This project reviews previous work in spatial robustness methods and proposes evolution strategies as zeroth order optimization algorithms to find the worst affine transforms for each input. The proposed method effectively yields robust models and allows introducing non-parametric adversarial perturbations.
\end{abstract}

\section{Introduction}
\label{sec:intro}
Adversarial attacks are the modifications, also known as perturbations in the literature, that transform an input with the purpose of causing the model to misclassify the modified inputs. Not any perturbation is allowed, but rather the modified inputs must be similar to the original input. For this reason, adversarial attacks specially play a key role in scenarios where reliability and security are important. The major work on robust training has been centered on perturbations constrained within an $\ell_p$ ball \cite{goodfellow, madry2017, free-adv, random-certified}. Even if these perturbations significantly drop the performance of state-of-the-art models, they are unlikely to exist in the real world. On the other hand, spatial transforms considered in this work can arise in real scenarios, e.g.\ by simply taking the same picture from a different position in the case of images. It has been shown that neural network-based image classifiers are also vulnerable to inputs modified with spatial transforms, in particular affine transforms \cite{fawzi}, which are the core of this project.

The idea of incorporating spatial transforms in the training of neural networks is not novel, since it was already studied in the context of data augmentation more than a decade ago. Data augmentation is based on randomly perturbing the samples from the training dataset and considering both the perturbed and the original sample in the training procedure. This has been considered as a measure to extend the cardinality of a dataset and to make the model invariant to the transformations considered in the process of augmentation. Moreover, data augmentation can act as a regularizer in preventing overfitting in neural networks \cite{data_aug1} and improve performance in imbalanced class problems \cite{data_aug2}. In fact, data augmentation is such a common practice that deep learning frameworks usually have built-in data augmentation utilities that not only allow to do affine transforms but more general projection transforms \cite{tensorflow}. 

Models trained with data augmentation offers relatively small robustness in front of adversarial attacks than those trained with the non-augmented dataset \cite{spatial}. In fact, in the presented experiments (see \Autoref{sec:exp}), they yield worse results than with standard training. Nevertheless, using carefully crafted perturbations during training instead of random perturbations before training as in data augmentation can improve the robustness of a classifier \cite{spatial}. This project focuses on implementing algorithms that can find adversarial examples with low computational cost on arbitrary transformations in order to perform robust training with them. Yet another goal is to accurately estimate the robustness against these arbitrary transformations.

\section{Problem formulation}
An adversarial example for an input tensor $\xx$ and a classifier $C$, is defined as a tensor $\xx'$ that causes $C$ to output a different label, i.e.\ $C(\xx)\neq C(\xx')$, while the adversarial example is similar (for a human) to the original input. In this work we will focus on the case where the inputs are images.

To the best of our knowledge, most of the previous works in adversarial robustness for images, uses error metrics that are based on pixel-wise signal differences. Concretely, they assume that $\xx'$ is a valid adversarial example if $\norm{\xx - \xx'}_p\leq \varepsilon$ for some $p\in [0,\infty]$, typically $p=\{1,2,\infty\}$, and a small enough $\varepsilon$ \cite{goodfellow, madry2017, free-adv, random-certified}.

Nevertheless, pixel-wise metrics do not reflect the similarity between an image and its version after applying an affine transform. Hence, small variations such as rotations and translations can lead to large values in all $p-$norms while being almost the same for a human. In fact, the concept of image similarity is very loosely defined in the adversarial robustness literature, since formally capturing the notion of human perception is extremely difficult \cite{visual_sim}. We address this issue in \Cref{sec:future}.

\begin{figure}[ht]
    \centering
    \begin{minipage}{.45\textwidth}
    \includegraphics[width=\textwidth]{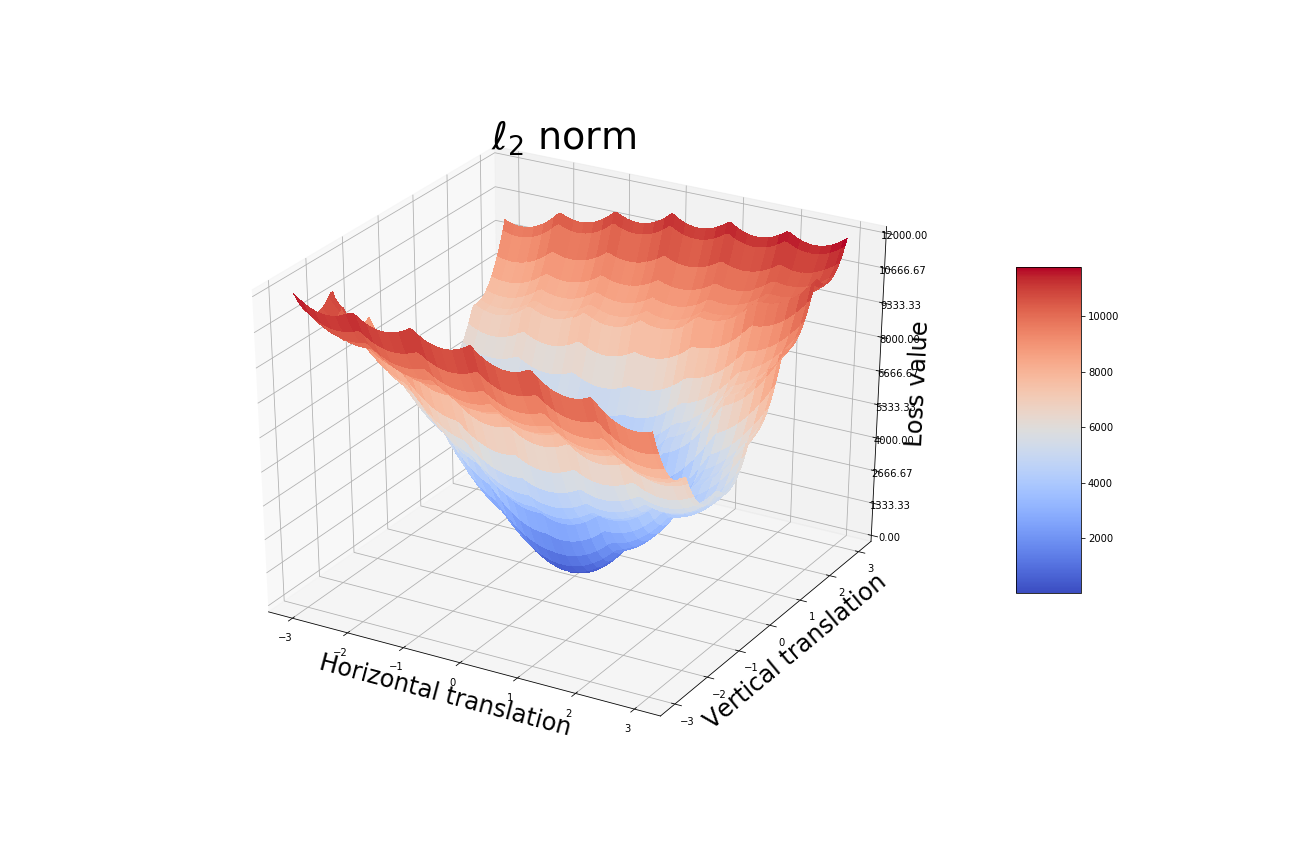}
    \end{minipage}
    \begin{minipage}{.45\textwidth}
    \includegraphics[width=\textwidth]{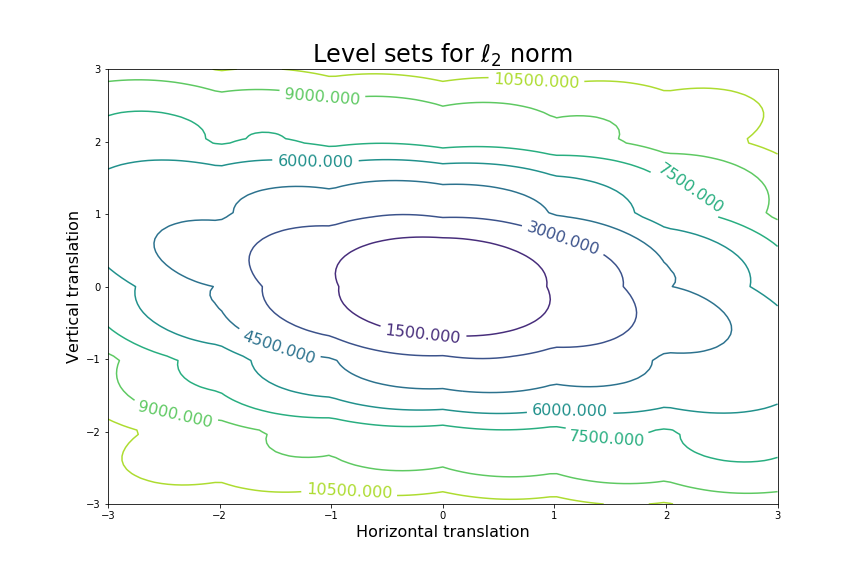}
    \end{minipage}
    \begin{minipage}{.45\textwidth}
    \includegraphics[width=\textwidth]{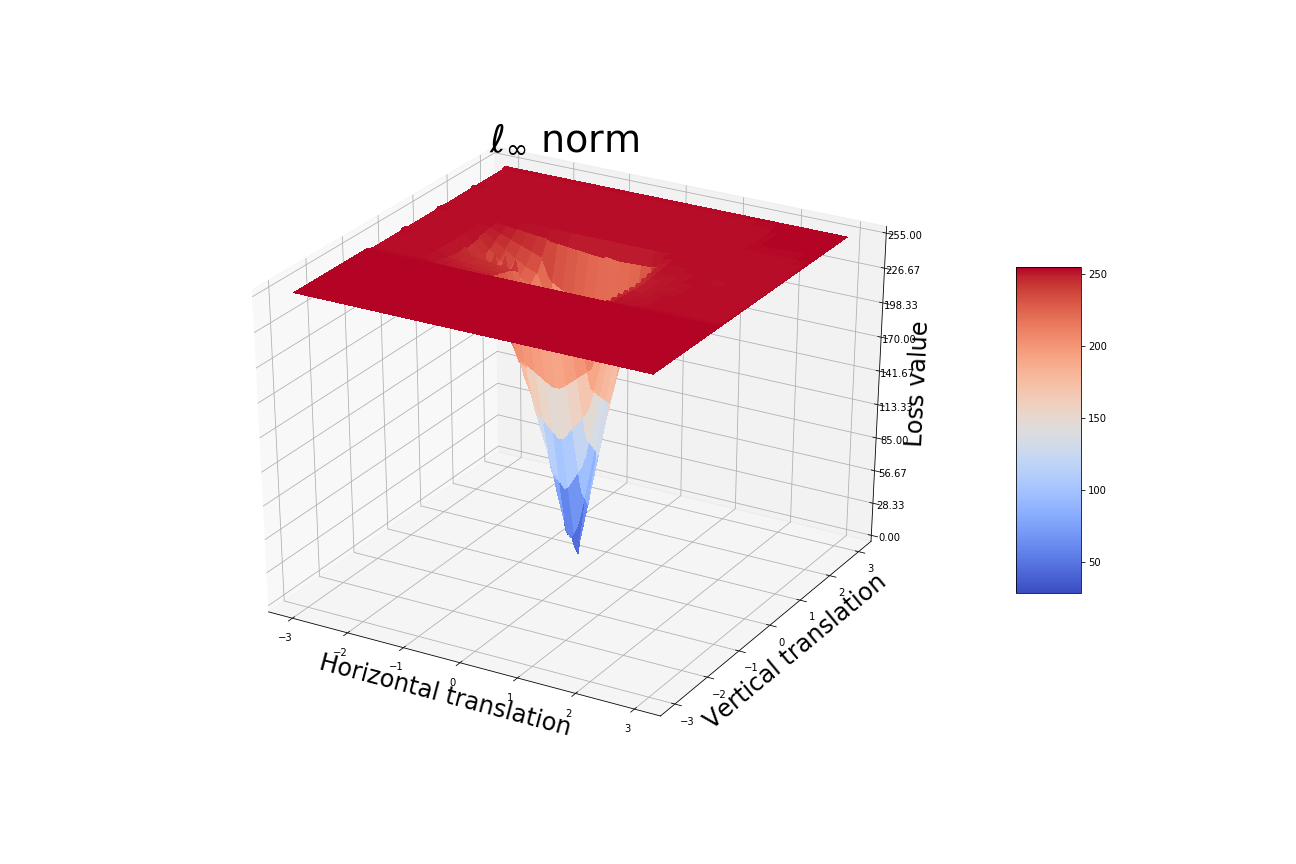}
    \end{minipage}
    \begin{minipage}{.45\textwidth}
    \includegraphics[width=\textwidth]{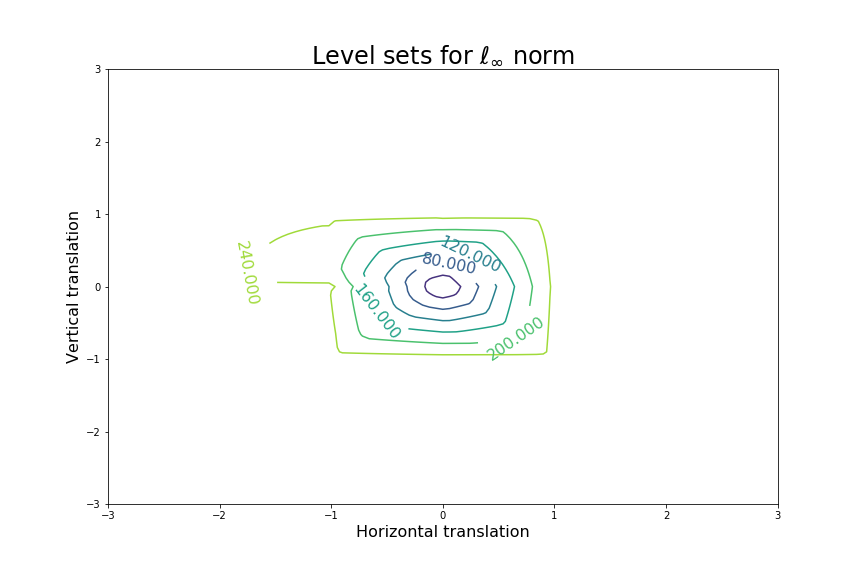}
    \end{minipage}
    \caption{$\ell_2$ and $\ell_\infty$ norms of difference signal $\xx-\xx'$ in the space of vertical and horizontal translations. This means that $\xx'$ is obtained by translating the image $\xx$ by the quantities (in pixels) of the x and y axis.}
    \label{fig:l2linf}
\end{figure}

\begin{figure}[ht]
    \centering
    \includegraphics[width=.4\textwidth]{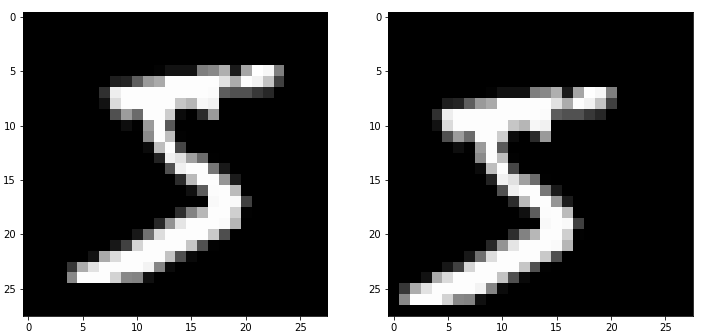}
    \caption{Original (left) and transformed (right) image with translation $(3,-2)$.}
    \label{fig:or_trans}
\end{figure}

\autoref{fig:l2linf} depicts the $\ell_2$ and $\ell_\infty$ norms when applying translations to a random image drawn from the MNIST dataset which can be seen in \autoref{fig:or_trans}. With the previous notation, that means $\xx'$ is the image resulting from translating $\xx$, and the graph plots $\norm{\xx-\xx'}_p$ for $p\in \{2,\infty\}$ against the horizontal and vertical translation in pixels.

An ideal metric for images, would be invariant against affine transforms, or at least for small translations (since as for a human, the images of \autoref{fig:or_trans} certainly represent the same number), and give a nearly flat loss landscape with almost zero value. This is not satisfied in \autoref{fig:landscape}. Moreover, $p-$norms are not necessarily symmetric in the space of translations, which would be another desirable property that can be easily seen to not hold for the $\ell_\infty$ case.

\section{Attack methods}
\label{sec:attack}
We extend the attack space described in \cite{spatial} by allowing general affine transforms, which include translations and rotations but also scale changes and shears. For a translation $(\delta_u, \delta_v)\in \R^2$, a rotation $\theta \in [0,2\pi)$, a shear $\phi \in [0,2\pi)$, and a scale $(s_u, s_v)\in \R_+^2$, the affine transform matrix for parameters $\mathbf{t}:=\begin{bmatrix}
    \delta_u &\delta_v&\theta&\phi&s_u&s_v
    \end{bmatrix}^T
    $ is defined as
\begin{align}
    A(\mathbf{t}) := \begin{bmatrix}
         s_u  \cos(\theta) & -s_v \sin(\theta + \phi)&  \delta_u \\
        s_u  \sin(\theta) &  s_v \cos(\theta + \phi)& \delta_v\\
        0 & 0 & 1 \\
    \end{bmatrix}
    \label{eq:trans}
\end{align}
where $\delta_u=\delta_v=\theta=\phi=0$ and $s_u=s_v=1$ recovers the identity matrix.

We formulate the robust training procedure as the following saddle point problem:
% Constrained space of possible perturbations yet to better define
\begin{align}
    \min_{W} \EEb{(\xx,y)\sim \cD}{\max_{\mathbf{t}\in \cS} \cL(W, f(\xx, A(\mathbf{t})), y)}
    \label{eq:problem}
\end{align}
where $W$ are the weights of the neural network, $\cL$ is the loss function, and $f:\R^{M\times N} \times \R^{3\times 3}\to \R^{M\times N}$ performs the affine transform to the input image $\xx\in\R^{M\times N}$. The function $f$ transforms the coordinates of a point $(u,v)$ on the original image to a new coordinate $(u',v')$ in the adversarial image as follows:
\begin{align}
    \begin{bmatrix}
        u'\\
        v'\\
        1
     \end{bmatrix}=A(\mathbf{t})\begin{bmatrix}
        u\\
        v\\
        1
     \end{bmatrix}
     \label{eq:map}
\end{align}
where $A(\mathbf{t})$ is the transformation matrix defined in \eqref{eq:trans}. Note that since $A(\mathbf{t})$ is a $3\times3$ matrix, input and output vectors are of dimension 3. Thus, the addition of a dummy third component is translated into a proper normalization of the coordinates of the transformed image. Put differently, $(u',v',w)$ represents a point at location $(u'/w,v'/w)$.

\begin{figure}[ht]
    \centering
    \begin{minipage}{.45\textwidth}
    \includegraphics[width=\textwidth]{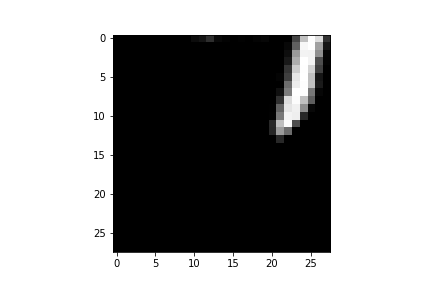}
    \end{minipage}
    \begin{minipage}{.45\textwidth}
    \includegraphics[width=\textwidth]{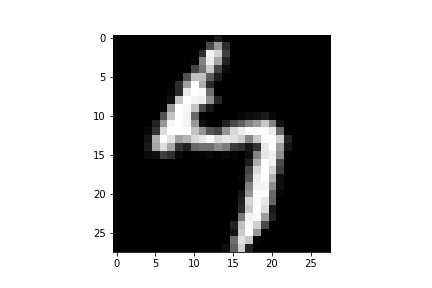}
    \end{minipage}
    \caption{Difference between centered and non-centered rotation (directly using \eqref{eq:trans}) with $\theta=50^\circ$.}
    \label{fig:centered}
\end{figure}

If a pixel gets mapped to a non-integer coordinate, i.e.\ $(u',v')\notin \mathbb{N}^2$, we compute its value by bilinear interpolation. Affine transforms sometimes require the value of pixels outside the image, hence padding is required. We use zero padding, which is specially meaningful for MNIST since the background and the padding are both black in this case (see \Autoref{fig:or_trans, fig:centered}).

The set $\cS \subseteq \R^d$ in \eqref{eq:problem} is a nonempty compact space of allowed perturbations containing the parameters associated to the identity operator. In this project, $\cS$ is a hyper-rectangle as the one defined in \cite{spatial}, that is, each parameter has handcrafted limits independent of other parameters as it can be seen in \autoref{tab:limits}.

\begin{table}[ht]
    \centering
    \begin{tabular}{cc}
        \toprule
        Parameter & Range \\
        \midrule
        $(\delta_u, \delta_v)$ & $[-3,3]^2$ \\
        $\theta$ (in degrees) & $[-30, 30]$ \\
        $\phi$ (in degrees) & $[-20, 20]$ \\
        $(s_u, s_v)$ & $[0.6, 1.4]^2$ \\
        \bottomrule
    \end{tabular}
    \caption{Parameter ranges for MNIST and CIFAR10.}
    \label{tab:limits}
\end{table}

Moreover, $\cS$ can be defined as a ball in some other metric as discussed in \Cref{sec:future}.

Sticking to the definition of adversarial examples, we want the transformed image to be similar to the original one. Directly applying the mapping \eqref{eq:map} can give an image that doesn't satisfy this similarity criterion as depicted in \autoref{fig:centered}. This is due to the implicit translation that a rotation, scaling and shearing imply. 

In order to avoid the latter, we enforce that the pixel in the center is mapped to its same position, i.e.\ we enforce that for $(u, v)=(\frac{M}{2}, \frac{N}{2})$, we have $(u', v')=(\frac{M}{2}, \frac{N}{2})$ in \eqref{eq:map}\footnote{Note that for odd M or N, we have to round $M/2$ or $N/2$ since the coordinates of a pixel are integral.}. Moreover, to simplify the re-centering, we sequentially apply translation, shearing, scaling and finally a rotation to the input image. When we say that we apply e.g.\ a translation, we mean that the only free parameters are $(\delta_u,\delta_v)$ and all the others are set to their value in the identity mapping.

The following sections introduce the baseline methods described in \cite{spatial} and propose two zeroth order optimization methods that find adversarial transformations.

\subsection{Baseline methods}
\label{sec:baseline}
The first baseline method is Grid search, which is based on discretizing $\cS$ and exhaustively examining every possible transformation in the discretized space. These transformations are evaluated sequentially, and for each image we pick the worst transformation according to \autoref{ass:worst_trans}. The Grid search method requires a lot of forward passes even when the number of free parameters is low. Therefore, this method is only used to test the robustness of trained models but not in the training procedure.

\begin{assumption}
    Let $\xx_j':=f(\xx,A(\mathbf{t}_j))$ be the image transformed with transform $\mathbf{t}_j$. When evaluating a set of transforms $\{\mathbf{t}_1,\dots,\mathbf{t}_i,\dots,\mathbf{t}_N\}$ for a given image $\xx$, transformation $\mathbf{t}_i$ is considered worse than the transformations $\{\mathbf{t}_1,\dots,\mathbf{t}_{i-1}\}$, if it either misclassifies the input while the previous transforms don't, i.e.\ $C(\xx_i')\neq y$, $C(\xx_j')= y$ $\forall j\in \{1,\dots,i-1\}$, or achieves the maximum loss in case of a tie. That is, when either $\textbf{t}_i$ doesn't misclassify neither do the previous ones, i.e.\ $C(\xx_j')= y$ $\forall j\in \{1,\dots,i\}$ but $i=\argmax_{j\in \{1,\dots, i\}} \cL(W,\xx_j',y)$, or $\textbf{t}_i$ misclassifies as some of the previous ones do and $i=\argmax_{j\in \{1,\dots, i\}:C(\xx_j')\neq y} \cL(W,\xx_j',y)$.
    \label{ass:worst_trans}
\end{assumption}

The second baseline method, which is used both in training and evaluation, is Worst-of-$k$. This method is based on sampling $k$ different transformations uniformly and independently sampled from $\cS$ and using the worst one (following \autoref{ass:worst_trans}) as an adversarial example. As we increase $k$, this attack interpolates between a random choice and Grid search.

Note that both baseline methods require specific numerical bounds for each parameter giving a feasible transform, i.e.\ we need to explicitly specify the support both to discretize and to define the uniform distribution over $\cS$. These methods have been adapted from \cite{spatial} and tested in the setup described in \Cref{sec:exp}.

\subsection{Evolution Strategies}
\begin{figure}[ht]
    \centering
    \begin{minipage}{.3\textwidth}
        \includegraphics[trim={8cm 2cm 10cm 3.5cm},clip,width=\textwidth]{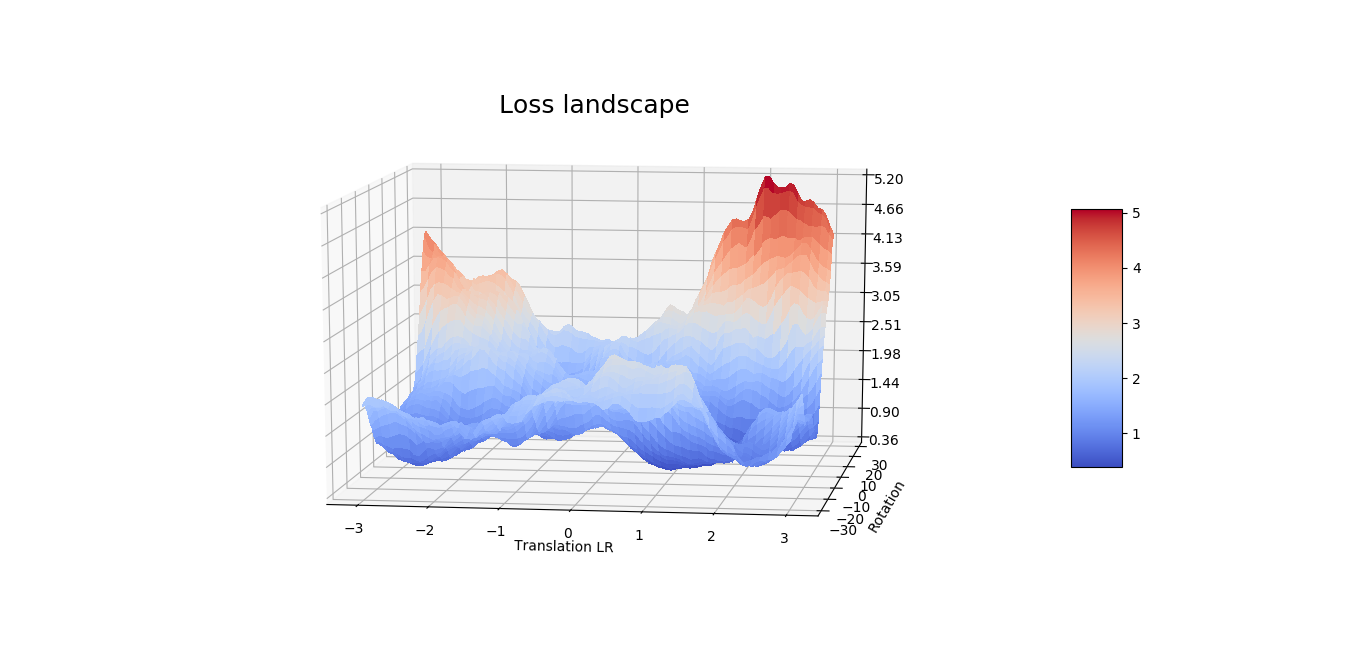}
    \end{minipage}
    \begin{minipage}{.3\textwidth}
        \includegraphics[trim={5cm 1cm 9cm 3cm},clip,width=\textwidth]{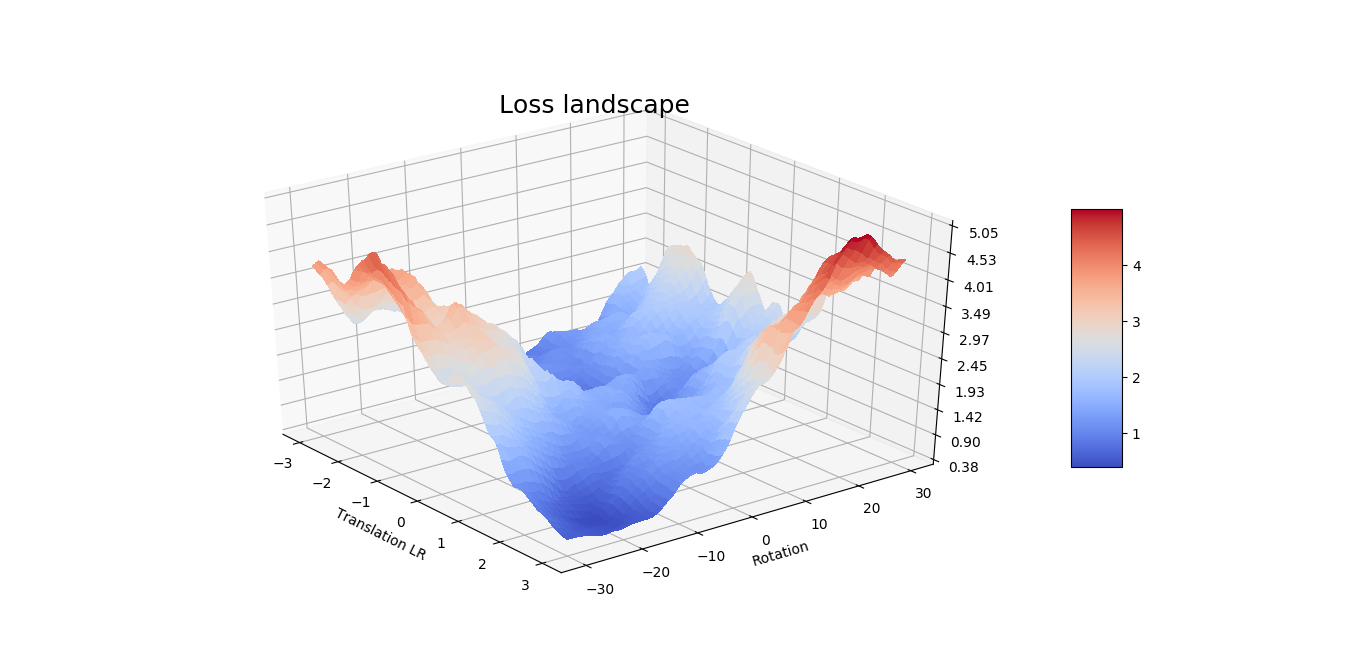}
    \end{minipage}
    \begin{minipage}{.3\textwidth}
        \includegraphics[trim={6cm 2cm 11cm 3cm},clip,width=\textwidth]{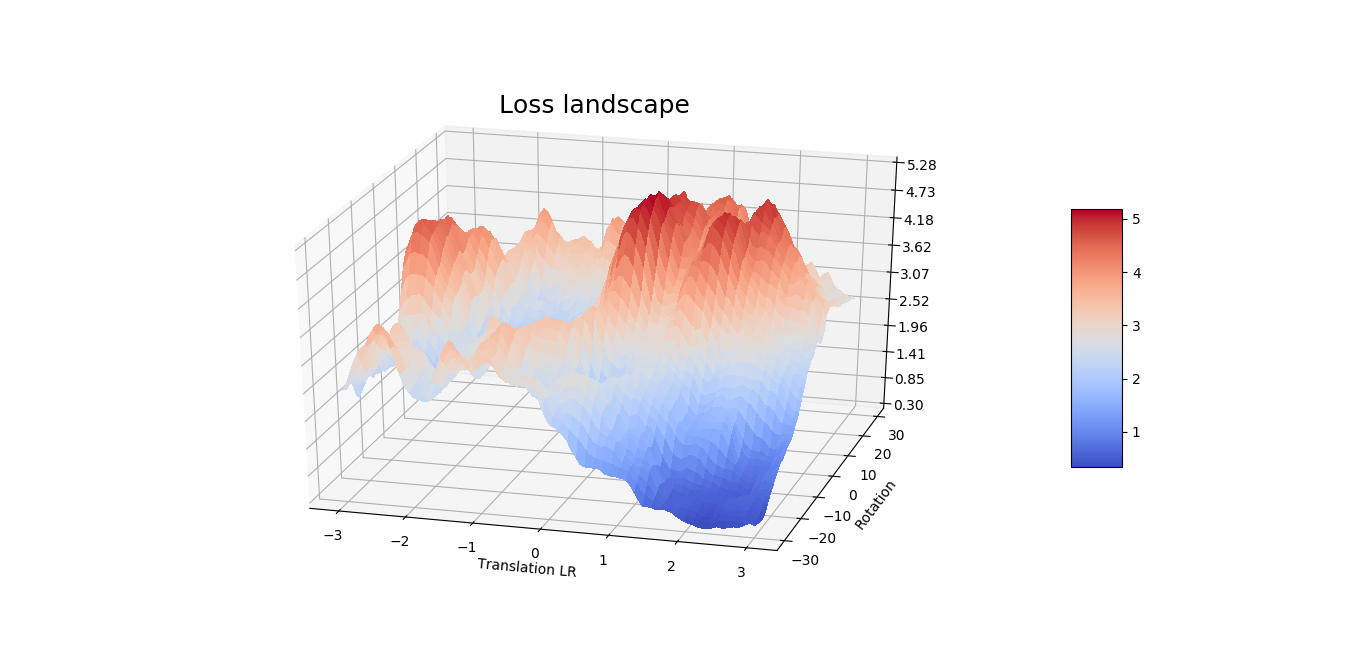}
    \end{minipage}
    \caption{Loss landscape of three random example drawn from the MNIST dataset when performing horizontal translations and rotations restricted according to \autoref{tab:limits}. The model and its loss function are described in \Cref{sec:exp}. In particular this model has been fully-trained using the standard training method.}
    \label{fig:landscape}
\end{figure}

As discussed in \cite{spatial}, due to the highly non-concave loss landscape in the space of the parameters defining the spatial transform, first-order methods lack any guarantee of optimality and cannot find good adversarial examples in practice. The loss landscape of some random samples from MNIST can be seen in \autoref{fig:landscape}.

\autoref{alg:es} provides the pseudo-code for a general evolution strategy adapted from \cite{cma} for constrained optimization. The two algorithms proposed in this section are variants that differ on how to update the previous parameters.

\begin{algorithm}[ht]
    \caption{\textsc{EvolutionStrategy$(F,\cS)$}}
    \label{alg:es}
    \begin{algorithmic}[1]
    \STATE \textbf{Input: }$F:\R^d \to \R$, the function to minimize and $\cS\subseteq \R^d$, the constraint set implicitly given by a membership oracle
    \STATE \textbf{Initialize: }Mean vector $\mm \in \cS$, step-size $\sigma\in \R^+$ and covariance matrix $C\in \R^{d\times d}$
    \WHILE{Stopping criterion not met}
        \REPEAT
            \STATE $\mathbf{t} \sim \mm + \sigma \cN(\0, C)$
        \UNTIL{$\mathbf{t} \notin \cS$, given by membership oracle}
        \STATE \textsc{Update}$(\mathbf{t}, \mm, \sigma, C, F)$
    \ENDWHILE
    \RETURN $\mathbf{t}$
    \end{algorithmic}
\end{algorithm}

In \autoref{alg:es}, we need to initialize $\mm\in \cS$, yet we claim that only a membership oracle is needed. Note that $\cS$ represents the set of allowed transformations, so one could initialize $\mm$ to be the parameters that give the identity operator. In case of having the aforementioned affine transform, this translates into setting $\mm = \begin{bmatrix}\delta_u& \delta_v& \theta&\phi&s_u&s_v\end{bmatrix}^T = \begin{bmatrix}0 & 0 & 0 & 0 & 1 & 1\end{bmatrix}^T$, which is contained in $\cS$. Following the notation in \eqref{eq:problem}, we can take $F: \mathbf{t} \mapsto -\cL(W, f(\xx, A(\mathbf{t})), y)$ for a given image $\xx$ with associated ground-truth label $y$, model weights $W$ and transformation $A(\mathbf{t})$. Nevertheless, in practice we choose the worst transforms following \autoref{ass:worst_trans}. This is because achieving a misclassification and a maximal loss are not the same even though both are related and the goal of an adversarial attack is to misclassify the transformed inputs. For the sake of notation, we represent this criteria with $F$. Note that reversing the sign is needed since \autoref{alg:es} requires a function to minimize and we are trying to find the transformation that maximizes the loss. Nevertheless, in the experiments we take $F$ such that the number of misclassified samples have more importance than the loss.

Unlike the sampling from a uniform distribution or the Grid search used in \cite{spatial} to find adversarial examples, given that the Gaussian has infinite support, we could get a transform that is not in the set of feasible transformations $\cS$. If this is the case, we resample the transformations such that $\mathbf{t} \notin \cS$, for which we only need the membership oracle. The number of times that we will have to resample depends on the definition of $\cS$. In the following section, we will further discuss the implications on the current setup.

In the current experiments, the stopping criterion of \autoref{alg:es} is simply a maximum number of iterations in the outer loop, which is equivalent to the number of calls to \textsc{Update}. As further explained in the next sections, in the case of (1+1)-ES, only one evaluation of $F(\mathbf{t})$ and hence one forward pass is needed to update the parameters. However, in the case of CMA-ES one needs several evaluations inside the \textsc{Update} function.

The key point of using evolution strategies is that, instead of naively sampling points in $\cS$, we can learn the worst transforms for each image separately at a given iteration. Thus, this can lead to a more robust model by incorporating the worst transforms during each training step, which by assumption change smoothly over the transformation space from one iteration to the next one. For this reason, the parameters $\mm,\sigma,C$ are stored for each image in the training set and when we call \autoref{alg:es}, we initialize $\mm,\sigma, C$ to the previous parameter values corresponding to the same image obtained on the previous epoch. The mean corresponds to the worst transform (according to \autoref{ass:worst_trans}) seen so far for each image. Note that in evaluation, \autoref{alg:es} is only run once per image and hence the parameters $\mm,\sigma,C$ are not stored for each image once the algorithm terminates.

\subsubsection{(1+1)-ES}
The simplest evolution strategy used to solve \eqref{eq:problem} is (1+1)-ES, which is based on sampling an isometric Gaussian with varying mean and variance. This means that in \autoref{alg:es}, we enforce $C=I$, the identity matrix. To effectively explore all the allowed transformation space $\cS$, we normalize all the parameters in such a way that its range is $[-1,1]^d$. Note that this needs $\cS$ to be symmetric around zero in every dimension, so we have to not only re-scale all the parameters but also shift the scale for each axis $(s_u,s_v)$ to be centered around 1 (see \autoref{tab:limits}). So in this case, we initialize $\mm=\0$.

Note that for the current setup, the probability of having the new sample in $\cS$ is at least $\frac{1}{2d}$, where $d$ is the number of free parameters of the transformation \eqref{eq:trans}, given that the previous iterate was in $\cS = [-1,1]^d$. In practice, we don't have to resample $\mathbf{t}$ too often given that the worst case is attained when sampling from a Gaussian centered in one corner of $\cS$. Moreover, the sampling of a Gaussian doesn't incur much overhead cost and the computation of the attack is dominated by the number of forward passes.

To avoid resampling, especially in the first iterates, we set the initial variance $\sigma_0$ such that the probability to resample in the first iterate, namely $\epsilon$ is low. We can analytically compute the initial sigma for a given $\epsilon$ as follows:

\begin{align}
    \label{eq:sigma0}
    \epsilon &:= \Pr[\mathbf{X} \notin [-1,1]^d]=\Pr[\lor_{i=1}^d X_i \notin [-1,1]] = 1 - \Pr[\land_{i=1}^d X_i \in [-1,1]] \\
    &= 1 - \Pr[X_1 \in [-1,1]]^d
    =1 - \left[ \int_{-1}^{1} \frac{1}{\sqrt{2\pi \sigma_0^2}} e^{-\frac{x^2}{2\sigma_0^2}} dx \right]^d
\end{align}
where $X_i \overset{\text{iid}}{\sim}\cN(0,\sigma_0^2)$. The choice of $\epsilon \approx 0.5$ gives the $\sigma_0=0.4$ that we use along the experiments with $d=3$. However, when using this method in evaluation, we found better results setting $\sigma_0=0.75$ since \autoref{alg:es} only runs for one times in this case. There are several variants of the (1+1)-ES algorithm. In this case, we follow the one-fifth success rule \cite{es}, resulting in the update described in \autoref{alg:up}. The variance increases if there is improvement with respect to the previous iteration and decreases for unsuccessful iterations. The factors of increase and decrease are theoretical to reach the one-fifth success rule described in \cite{es}.

\begin{algorithm}[ht]
    \caption{\textsc{Update$(\mathbf{t}, \mm, \sigma, C, F)$} for (1+1)-ES with 1/5-th success rule}
    \label{alg:up}
    \begin{algorithmic}[1]
    \STATE \textbf{Input: }Transform $\mathbf{t}$, mean $\mm$, step-size $\sigma$, Covariance $C$, and $F:\R^d \to \R$, the function to minimize
    \IF{$F(\mathbf{t})\leq F(\mm)$}
        \STATE $\mm \leftarrow \mathbf{t}$
        \STATE $\sigma \leftarrow 1.5\sigma$
    \ELSE
        \STATE $\sigma \leftarrow 1.5^{-1/4}\sigma$
    \ENDIF
    \end{algorithmic}
\end{algorithm}

Given the nature of \autoref{alg:up} along with the memory of the evolution strategy in training, $\sigma$ may vanish. Empirically, we observed that indeed the step-size tends to vanish and even if theoretically possible, it never explodes. When $\sigma\approx0$ we get a deterministic behavior, i.e.\ $\mathbf{t}$ is no longer updated and takes the value $\mathbf{t}\approx \mm$. This happens for example when $\mm$ corresponds to a local maximum. In this case, all the transformations that are close to $\mm$, which are the ones sampled with higher probability are worst in terms of the objective value. Thus in such situation it is virtually impossible to improve the current solution if $\sigma$ is small. To tackle this, we introduce a restart condition that sets the variance back to its initial value $\sigma_0$. This restart condition is based on comparing the current variance to the equivalent of having 5 decreases in a row from the initial iteration. If the current $\sigma$ is lower than this threshold, we restart it. This translates into adding
\begin{align}
    \sigma \leftarrow 
    \begin{cases}
        \sigma_0 &\text{if }\sigma\leq \sigma_0 (1.5^{-1/4})^5\\
        \sigma &\text{otherwise}
    \end{cases}
    \label{eq:restart}
\end{align}
at the end of \autoref{alg:up} for (1+1)-ES.

\subsubsection{CMA-ES with diagonal decoding and unitary determinant}
\label{sec:cma_es}
Especially for larger values of $d$, the number of free parameters in \eqref{eq:trans}, it is interesting to learn relations among parameters. In other words, it could be that there is a correlation between parameters that yield a large loss and the independence assumption of (1+1)-ES would be to simplistic in this case.

In order to see if indeed the (1+1)-ES algorithm was too simple, we tested CMA-ES, which allows full covariance matrices \cite{cma} and not just a fixed identity matrix. In particular we use CMA-ES with diagonal decoding. which defines the \textsc{Update} function used in this case \cite{ddcma}.

Avoiding a deterministic behavior is not as easy as comparing a scalar with a threshold as in \eqref{eq:restart} for the case of (1+1)-ES. The determinant, which can be interpreted as the oriented volume of the hyper-parallelepiped defined by the column vectors of $C$, can be used to spot such deterministic behavior. This latter is associated to the case when the determinant, and hence the volume in the former interpretation, is too small.

Instead of restarting the covariance matrix, we fix our determinant to be unitary, which translates into adding the step
\begin{align}
    C \leftarrow \frac{C}{\det(C)}
    \label{eq:restart_cov}
\end{align}
at the end of the \textsc{Update} procedure described in \cite{ddcma} to prevent shrinking.

Even if in this case there is no need to normalize the parameters since the Gaussian in CMA-ES is not necessarily isometric, it is simpler to do so in the case of $\cS$ being an hyper-rectangle. This simplifies the implementation of the membership oracle and allows using the same initialization as for the (1+1)-ES algorithm. The caveat of this algorithm, is that in order to update the full covariance matrix, we need to perform several evaluations of the objective function with the same parameters. Such number of evaluations with fixed parameters is known as the population size. In particular, this population size has to be of at least 3, and by default it is set to $4 + \lfloor 3 \log(d) \rfloor$ \cite{ddcma}. If comparing algorithms with respect to the number of forward passes, one should take this into account and reduce the number of iterations in the outer loop of \autoref{alg:es}, which is the stopping criterion of the latter.

As stated before, we use this method for larger values of $d$, implying that the probability of resampling is higher. Differing from (1+1)-ES, we don't have independence of coordinates. Moreover, the second equality of \eqref{eq:sigma0} doesn't hold in this case due to the dependence of coordinates. Thus, it could happen that given the value in some coordinates in $\cS$, it is very unlikely to get feasible values, i.e.\ $\Pr[\land_{i\in \cX} X_i\in [-1,1]| \land_{i\notin \cX} X_{i}\in [-1,1]]$ may be very low for some $\cX\subseteq [d]$. 

To avoid resampling, we introduced two variants that take advantage of the normalization of the transform space into the hyper-cube $\cS=[-1,1]^d$. These are projection onto $\cS$ and applying $\tanh(\cdot)$ element-wise to the sampled transformation \cite{tanh}. Applying the hyperbolic tangent to the samples of the Gaussian is compatible with the infinite support of such distribution yet avoids concentration in the extreme points as with the projection approach.

\section{Experiments}
\label{sec:exp}
The set of allowed transformations ($\cS$ as defined in \eqref{eq:problem}) in all the presented experiments is defined as the Cartesian product of all the sets of \autoref{tab:limits}. Note that in the latter, $\cS$ is defined for specific datasets, which is because parameters such as the translation and the scaling are more meaningful if they depend on the image size or the nature of the dataset. We only allow a translation of $10\%$ of the image size\footnote{MNIST images are of size $M=N=28$ pixels and CIFAR10 images of size $M=N=32$ pixels. After rounding to the nearest integer we get the $\pm3$ of \autoref{tab:limits}.} as in \cite{spatial}. The other limits were chosen so as to give visually similar transformed images in the limit cases. Put differently, even in limit cases, it should be clear for a human that the label of the original and transformed images are the same.

\textbf{Model architecture: }For MNIST, we use a convolutional neural network consisting of two convolutional layers with 32 and 64 filters each. Both layers are followed by a max pooling layer. At the end of the last convolutional layer, two fully connected layers follow. ReLU is chosen as activation function for all cases, and Cross Entropy is used as the loss function $\cL$. This model corresponds to the CNN of the MNIST Challenge \cite{madry2017}, which is different from the model for the same dataset used in \cite{spatial}. This means that the results of the experiments won't necessarily be the same even though the Worst-of-$k$ and Grid search methods are virtually the same as in \cite{spatial}. We used a different model since the description of the MNIST model used in \cite{spatial} only explains the differences with respect to a baseline model given in the TensorFlow tutorial \cite{tensorflow}, which is no longer existent. For all methods, we used 10k training iterations, that is roughly 21 epochs.

\textbf{Data augmentation: }The data augmentation procedure used in the following experiments is the same used in \cite{spatial}. This consists in doing random flips in the horizontal dimension of an image. In particular, with probability 0.5 we get the same image and with probability 0.5 we get the flipped version. Note that a flip in the horizontal dimension can be modelled as an affine transform with $\theta=\phi=\pi$. Clearly this transform is not included in $\cS$ (see \autoref{tab:limits}) yet it yields more robust models in \cite{spatial}. As a further clarification, adversarial perturbations are computed at each training iteration and the transformed versions of the inputs are used as the training dataset for such iteration. Nevertheless, for this implementation of data augmentation, perturbations are random and computed once before training starts. Such transforms and the original inputs are used as the training dataset, which remains static in all the training procedure.

\autoref{tab:exp_1} shows the accuracy of different classifiers on MNIST with affine transforms restricted to only rotation and translation (both vertical and horizontal), hence $d=3$. Note that the Grid search method is clearly the most effective one\footnote{When evaluating adversarial attacks, the lower the accuracy in evaluation, the better. When the attack is incorporated in training, we expect the accuracy drop to be significantly damped with respect to (at least) standard training.}, but with the current setup and the granularity described in \cite{spatial}, it takes 775 forward passes to find each attack. Thus, it is infeasible to incorporate this method in training, even with only 3 dimensions. When $d$ increases, the number of forward passes for Grid search explodes, since it is $\cO(\exp(d))$.

\begin{table}[ht]
    \centering
    \caption{Accuracy averaged over 3 classifiers on MNIST trained with different seeds. The rows and the columns represent the training and evaluation methods respectively. For each pair of training method and evaluation, the mean and standard deviation of the accuracies is provided. The best accuracies for each attack and the overall best methods for robust training and attacking are in bold font.}
    \begin{tabular}{c|ccccc}
      \toprule
      Model & Nat. & W-10 & \textbf{Grid} & (1+1)-ES-10 & (1+1)-ES-100\\
      \midrule
      Standard & $\boldsymbol{99.05\pm 0.06}$ & $27.19\pm 0.15$ & $0.00\pm 0.00$ & $37.61\pm 0.40$ & $37.87\pm 0.52$\\
      Aug. 30 & $98.40\pm 0.13$ & $14.18\pm 0.67$ & $0.00\pm0.00$ & $21.93\pm0.80$ & $21.55\pm0.90$\\
      W-10 & $98.91\pm 0.00$ & $\boldsymbol{96.81\pm0.01}$ & $87.16\pm 0.00$ & $96.82\pm 0.03$ & $96.82\pm 0.03$ \\
      (1+1)-ES-10 & $98.88\pm 0.00$ & $95.97\pm0.05$ & $86.24\pm0.00$ & $96.09\pm 0.05$ & $96.08\pm 0.07$ \\
      (1+1)-ES-20 & $98.96\pm 0.02$ & $96.77\pm0.01 $ & $87.05\pm 0.00$ & $96.17\pm 0.05$ & $96.17\pm 0.05$\\
      \textbf{(1+1)-ES-50} & $98.92\pm 0.08$ & $96.76\pm 0.07$ & $\boldsymbol{90.08\pm0.00}$ & $\boldsymbol{96.97\pm 0.06}$ & $\boldsymbol{96.97\pm 0.00}$\\
      \bottomrule
    \end{tabular}
    \label{tab:exp_1}
\end{table}

As seen in \autoref{tab:exp_1}, the evaluation with Grid search always yields lower accuracy than the other methods, thus being the best procedure to find adversarial examples for trained models. Nevertheless, even evaluating with Grid search can take a long time and thus other methods of evaluation can be preferred in some setups. Also note than Worst-of-10 can find adversarial examples yielding lower accuracies than (1+1)-ES for most of the models. This is because in this case, the key point of learning the parameters for each individual image along several epochs cannot be used. Furthermore, in this case having to sample from Gaussians may impair that adversarial examples yielding lower accuracies in classification, usually located near the extreme points of $\cS$, are explored.

Given that the (1+1)-ES algorithm only updates the transform if it is better, again regarding the same criterion described in \autoref{ass:worst_trans}, adding (1+1)-ES iterations can only improve. However, when evaluating with (1+1)-ES, only mild improvements are observed when augmenting the number of forward passes from 10 to 100.

Overall, one can observe that data augmentation is the worst approach for robustness in all cases, yielding even worse results than with standard training. This is attributed to the fact that the transformation used in data augmentation is not representative of the adversarial examples seen when evaluating. We tried this method of data augmentation since it yielded better results than standard training in \cite{spatial} even if the transform is also not in the attack space. Intuitively, incorporating the same affine transforms in data augmentation should increase the robustness of the model, but we wanted to take the setting of \cite{spatial} as a baseline.

W-10 and (1+1)-ES obtain very similar accuracies, and in particular the variant with 50 iterations, (1+1)-ES-50, achieves better results according to the strongest attack (Grid search) for $d=3$. Nonetheless, having 5 times as many forward passes per training iteration doesn't increase much the accuracy in general as seen in \autoref{tab:exp_1}, but it roughly multiplies the training time by a factor of 5. Thus, in some setups, alternatives with less forward passes may be preferred. Finally, the best performance with natural evaluation was achieved by the standard model in both experiments as expected. Thus, in comparison to standard models, robust training yields models that have slightly worst accuracy for the test dataset but much better accuracy for the same inputs after being adversarially transformed.

\autoref{tab:exp_2} shows the accuracies for different models on MNIST in the case of general affine transforms. As aforementioned, the focus of this project is building robust models rather than proposing attack methods. Moreover the best attack, as seen in \autoref{tab:exp_1} can be achieved by using Grid Search and, as seen in \autoref{tab:exp_1}, ES-(1+1) is not better than Worst-of-$k$ in evaluation. Thus, in \autoref{tab:exp_2} only natural and Worst-of-$k$ evaluations are provided.

Regarding the CMA-ES method, the hypothesis of a possible non termination of the inner loop of \autoref{alg:es} discussed in \Cref{sec:cma_es} is confirmed. Indeed, in all the cases it happened that eventually, given the values of some parameters of the transform, it is nearly impossible to get feasible values for the rest of parameters. Therefore, the experiments for CMA-ES were only conducted for the projected and the hyperbolic tangent approaches. For this methods we use 12 forward passes, that is 2 calls to \textsc{Update} for each training iteration. This is because in this case the population size is $4+\lfloor3\log 6\rfloor = 6$. In \autoref{tab:exp_2}, we also test whether the determinant normalization \eqref{eq:restart_cov} helps or not. The suffix FreeDet in the models of \autoref{tab:exp_2} means that the determinant is free in such case, hence no determinant normalization.

\begin{table}[ht]
    \centering
    \caption{Accuracy averaged over 3 classifiers on MNIST trained with different seeds along with its standard deviation. In this case, we allow general affine transformations, that is $d=6$.}
    \begin{tabular}{c|cccc}
      \toprule
      Model & Nat. & W-10 & \textbf{W-100}\\
      \midrule
      Standard & $\mathbf{98.91\pm 0.00}$ & $0.16\pm0.00$ & $0.00\pm0.00$ \\
      Augmented & $98.61\pm 0.04$ & $0.06\pm 0.01$ & $0.00\pm0.00$\\
      \textbf{W-10} & $97.05\pm0.00$ & $\mathbf{75.91\pm0.00}$ & $\mathbf{39.58\pm0.00}$ \\
      (1+1)-ES-10 & $92.29\pm 9.95$ & $54.03\pm 30.10$ & $24.14\pm15.62$\\
      CMA-ES-12-Proj & $9.80\pm0.00$ & $9.80\pm0.00$ & $9.80\pm0.00$ \\
      CMA-ES-12-Tanh & $9.80\pm0.00$ & $9.80\pm0.00$ & $9.80\pm0.00$ \\
      CMA-ES-12-Proj-FreeDet & $34.48\pm0.00$ & $0.30\pm0.00$ & $0.00\pm0.00$\\
      CMA-ES-12-Tanh-FreeDet & $31.87\pm0.00$ & $0.59\pm0.00$ & $0.00\pm0.00$\\
      \bottomrule
    \end{tabular}
    \label{tab:exp_2}
\end{table}

As seen in \autoref{tab:exp_2}, for $d=6$ the seed has less influence in all cases but (1+1)-ES-10. In this latter, we can see huge variance on the results. In the other cases, there is barely randomness in the accuracies of the models trained with the same procedure and different seeds. Even though the approaches deriving from CMA-ES-12 are promising in theory, the results are worst than those achieved with $C=I$ as in (1+1)-ES-10. This is probably due to the non-convergence of the covariance matrix. Empirically, we observed that the spectral norm of the difference of consecutive covariance matrices is not decreasing as it would be expected. Thus, the approaches of CMA-ES described in \Cref{sec:cma_es} are not a good alternative for efficiently training robust models. This alternative would probably improve if updating the covariance matrix for more than just 2 times with fixed parameters. Nevertheless, at the cost of having more forward passes, one could use Worst-of-$k$ which would approach the grid search performance as $k$ increases (see \Cref{sec:baseline}).

The best accuracies for affine adversarial attacks are obtained with the model trained with W-10. In 2 out of the 3 models trained with (1+1)-ES the results were very similar to those obtained with W-10. In particular, the accuracies were in both cases 98.04, 71.41 and 33.16 for natural, W-10 and W-100 evaluation respectively. Nevertheless, the results with the third seed were much worse, which shows the instability of this method with $d=6$. Thus in this case W-10 would clearly be the preferred method for robust training.

\section{Conclusions and future work}
\label{sec:future}
To sum up, the proposed methods allow to effectively train robust models against affine transform attacks. The initialization, handling of deterministic behavior and update strategy gives room for the improvement in the use of \autoref{alg:es} for finding better adversarial examples.

Other more complex adversarial attacks such as JPEG compression with different quality levels such as in \cite{jpeg_attack} could be constrained within a ball defined using the Structural Similarity Index (SSIM) described in \cite{ssim} as the metric to determine which inputs are valid adversarial examples. This metric was actually conceived in the image processing literature taking into account the human visual system and aiming to measure the image quality based on an initial uncompressed or distortion-free image as reference. Hence, it is more meaningful to define which adversarial examples do not change the semantics of the image instead of the definitions using $\ell_p$ norms or handcrafted limits as in \cite{spatial}.

In general, the possibility of implicitly give $\cS$ by a membership oracle for (1+1)-ES allows using more complex metrics and looking for affine-invariant alternatives that give meaningful transforms for a ball of fixed radius. Using an implicit definition of $\cS$ may also be possible for CMA-ES but the presented formulation failed to terminate in practice. In order to tackle with the infinite resampling, we introduced two variants that don't need to resample and thus can compute adversarial examples at a speed comparable to the Worst-of-$k$ method with $k$ being the same number of forward passes. Nevertheless this approaches are not compatible with $\cS$ given by a membership oracle, but need a projection operator in the projected version, or a constrained set described by an hyper-cube in the variant with hyperbolic tangent.

\section{Acknowledgments}
I would like to express my sincere gratitude to my advisor Dr.~Sebastian Stich for introducing me to the marvelous field of adversarial robustness and for its dedication and continuous support.

I am also particularly grateful for the assistance given by Maksym Andriushchenko. He brought us back to earth by providing a paper implementing what we thought was a great novel idea \cite{spatial}, and he also provided his broad knowledge in adversarial robustness.

Last by not the least, I would like to thank Prof.~Martin Jaggi for his valuable feedback.

\clearpage
%\bibliography{xxx} %add your bibliography
\bibliography{references}
\bibliographystyle{myplainnat}

\newpage
\onecolumn
\appendix

\end{document}

%% file: epfml_macros.tex
\usepackage{amssymb,amsmath,amsthm,dsfont}

\providecommand{\norm}[1]{\ensuremath{\left\lVert#1\right\rVert}}

  \providecommand{\R}{\mathbb{R}} % Reals
  \DeclareMathOperator{\E}{{\mathbb E}}
  \providecommand{\EEb}[2]{\ensuremath{\E_{#1}\!\! \left[#2\right] }} %expectation,  with brackets

  \DeclareMathOperator*{\argmax}{arg\,max}

  \providecommand{\0}{\mathbf{0}}

  \providecommand{\mm}{\mathbf{m}}

  \providecommand{\xx}{\mathbf{x}}

  \providecommand{\cD}{\mathcal{D}}

  \providecommand{\cL}{\mathcal{L}}
  
  \providecommand{\cN}{\mathcal{N}}
  \providecommand{\cO}{\mathcal{O}}

  \providecommand{\cS}{\mathcal{S}}

  \providecommand{\cX}{\mathcal{X}}

\RequirePackage[colorinlistoftodos,bordercolor=orange,backgroundcolor=orange!20,linecolor=orange,textsize=scriptsize]{todonotes}
\providecommand{\comment}[2]{\todo[inline,caption={}]{\textbf{#1: }#2}}%
\providecommand{\inlinecomment}[3]{%
  {\color{#1}#2: #3}}%
\newcommand\commenter[2]%
{%
  \expandafter\newcommand\csname i#1\endcsname[1]{\inlinecomment{#2}{#1}{##1}}
  \expandafter\newcommand\csname #1\endcsname[1]{\comment{#1}{##1}}
}

\newtheorem{assumption}{Assumption}

\definecolor{mydarkblue}{rgb}{0,0.08,0.45}